\documentclass{article}

\PassOptionsToPackage{numbers, compress}{natbib}

    \usepackage[final]{aloe_neurips_2023}

\usepackage[utf8]{inputenc} 
\usepackage[T1]{fontenc}    

\usepackage{url}            
\usepackage{booktabs}       
\usepackage{amsfonts}       
\usepackage{nicefrac}       
\usepackage{microtype}      
\usepackage{xcolor}         
\usepackage{tabularray}
\usepackage{enumitem}
\usepackage{graphicx}
\usepackage{caption}
\usepackage{subcaption}
\usepackage{microtype}
\usepackage{booktabs}
\usepackage{hyperref}
\usepackage{pgfplots}
\usepackage{amsmath}
\usepackage[capitalize,noabbrev]{cleveref}
\usepackage{wrapfig}
\usepackage{tcolorbox}
\usepackage{xspace}
\usepackage{natbib}
\usepackage{bbm}

\hypersetup{
    colorlinks=true,
    linkcolor=orange,
    filecolor=magenta,      
    urlcolor=orange,
    citecolor=orange,
}

\newcommand{\our}{LiFT\xspace}
\newcommand{\ourlong}{LiFT (unsupervised \textbf{L}earning w\textbf{i}th \textbf{F}oundation models as \textbf{T}eachers)}

\newcommand{\multilinecell}[2][c]{%
  \begin{tabular}[#1]{@{}c@{}}#2\end{tabular}}

\newcommand{\iconsheep}{\includegraphics[height=1.2\fontcharht\font`\B]{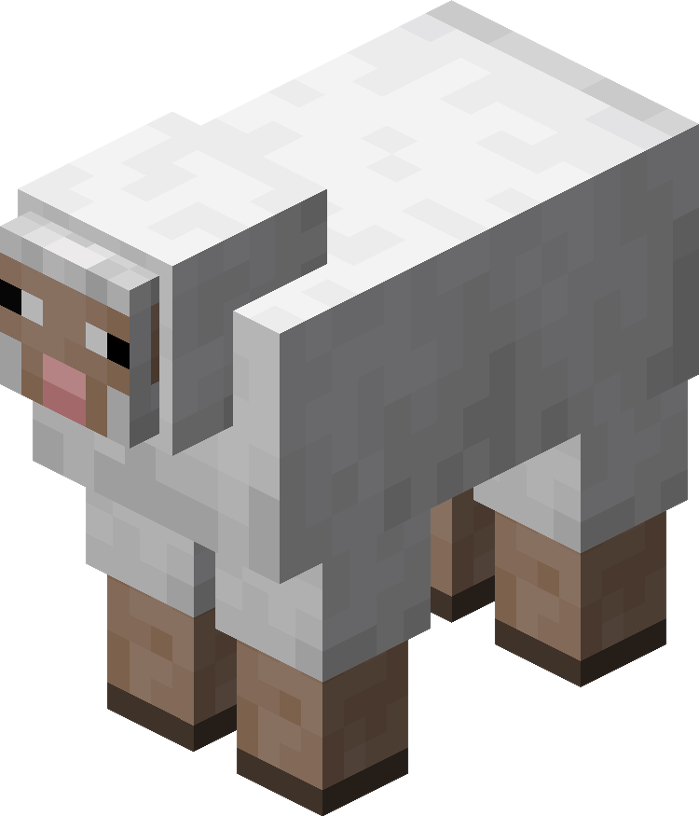}\xspace}
\newcommand{\iconshear}{\includegraphics[height=1.2\fontcharht\font`\B]{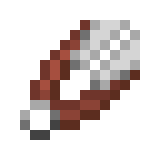}\xspace}

\title{
\our: Unsupervised Reinforcement Learning
\\ with Foundation Models as Teachers
}

\author{%
Taewook Nam$^{1}$\thanks{Equal contribution.} \quad Juyong Lee$^{1*}$ \quad Jesse Zhang$^2$ \\ \quad \textbf{Sung Ju Hwang}$^1$ 
\quad \textbf{Joseph J. Lim}$^1$ \quad \textbf{Karl Pertsch}$^{3,4}$\\
$^1$KAIST $^2$University of Southern California\\
$^3$University of California, Berkeley $^4$Stanford University\\
\texttt{\{namsan,agi.is\}@kaist.ac.kr}\\
}

\begin{document}
    \maketitle

    \begin{abstract}
We propose a framework that leverages foundation models as teachers, guiding a reinforcement learning agent to acquire semantically meaningful behavior without human feedback. 
In our framework, the agent receives task instructions grounded in a training environment from large language models. 
Then, a vision-language model guides the agent in learning the multi-task language-conditioned policy by providing reward feedback.
We demonstrate that our method can learn semantically meaningful skills in a challenging open-ended MineDojo environment while prior unsupervised skill discovery methods struggle. 
Additionally, we discuss observed challenges of using off-the-shelf foundation models as teachers and our efforts to address them.
\end{abstract}
    \section{Introduction}

A longstanding goal of robot learning and reinforcement learning (RL) is to train general agents that can perform a wide variety of helpful tasks in many different environments.
Recently, researchers have used imitation learning and RL to train such general agents~\cite{lynch2019play, pertsch2021skild, ebert2021bridge, heo2023furniturebench, zhang2023sprint, rt22023arxiv, chebotar2023qtransformer}, but these methods generally require costly supervision in the form of large-scale human demonstration datasets, which is hard to scale.
Unsupervised RL attempts to mitigate this issue by equipping agents with diverse skills through unsupervised learning objectives. 
Prior work generally teaches agents skills by optimizing information-theoretic diversity measures~\cite{DIAYN,APT,CIC,MaxEnt,DADS,LSD}, but the learned skills are often meaningless for downstream tasks as diversity measures do not induce \emph{meaningful} skills.

Imagine if instead, an agent deployed in a new environment had a human teacher, e.g., a robot dropped into a hospital with medical staff teaching them to perform important hospital duties. 
Not only would the teacher be able to guide the agent towards meaningful tasks to learn (e.g., finding certain medicines, checking on patients, etc.), but also the teacher could give feedback about how well the agent performed the task.
This is a clear intuitive way of teaching agents, but the required supervision is too costly to scale.
Can we teach RL agents in this manner but \emph{without} human supervision?

Our key insight is to use foundation models (FMs) as teachers; by leveraging the common sense knowledge captured in FMs through web-scale pre-training, the agent can learn semantically meaningful behaviors, similar to how human teachers can steer the agent’s learning process.
Prior works have indeed demonstrated that the knowledge captured by FMs is useful for recognizing diverse objects~\citep{gu2022openvocabulary}, executable tasks~\cite{ELLM, Voyager, OMNI, zhang2023bootstrap} and task successes~\cite{ZEST, MineDojo, escontrela2023video, LAMP}. In this work, we leverage FMs to design an unsupervised RL approach to learn complex visuomotor tasks in lieu of \emph{any} human supervision over what tasks to learn \emph{or} how to reward the agent for those tasks.

\begin{figure}[t]
    \centering
    \includegraphics[width=\textwidth]{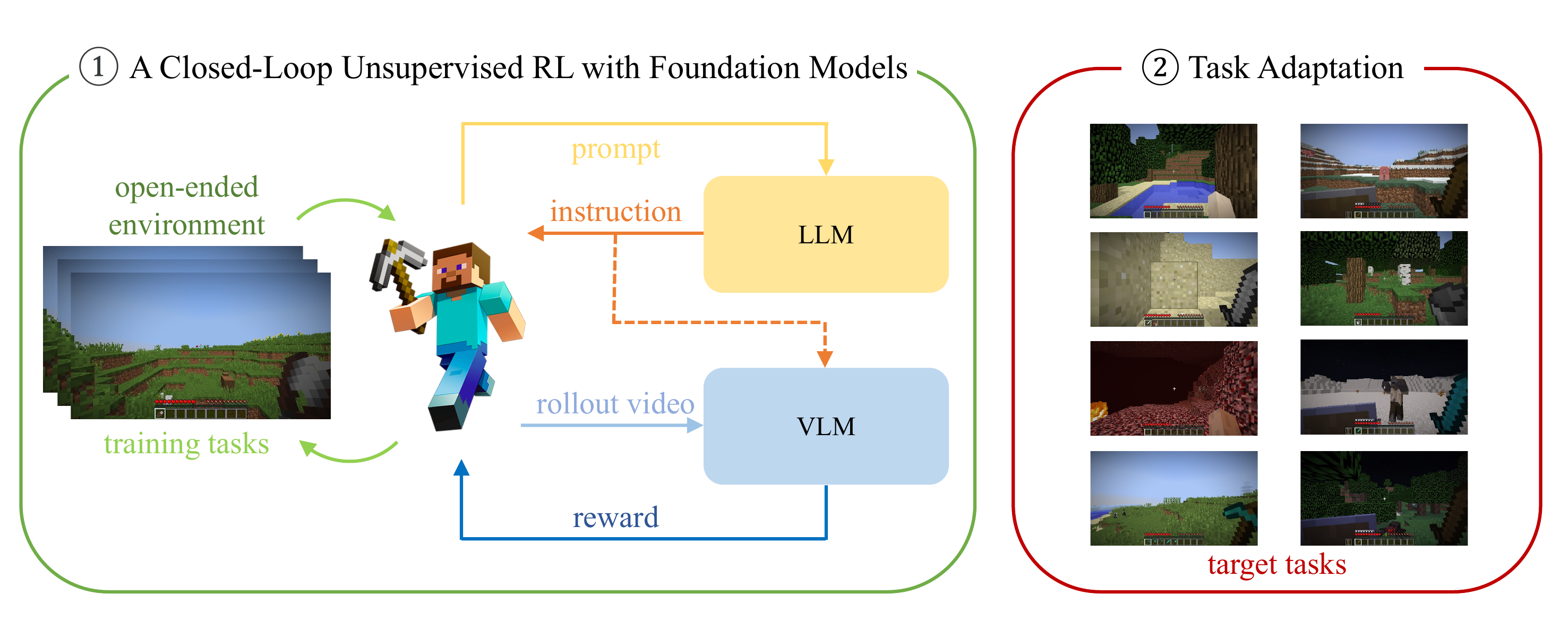}
    \caption{\textbf{\our overview.} Foundation models teach an agent in training environments by guiding them towards which tasks to learn (\textcolor{yellow}{LLM}) and rewarding them (\textcolor{cyan}{VLM}) for completing the tasks. The agent learns meaningful skills from the training environment and adapt to evaluation task instructions.}
    \label{fig:overview}
\end{figure}



Specifically, we propose \ourlong, an unsupervised RL framework that can acquire meaningful behaviors in challenging open-ended environments \emph{without} human supervision.
Concretely, we leverage large language models (LLMs) to propose meaningful task instructions for a language-conditioned agent to attempt given a list of objects in the scene. 
After the agent attempts to solve the task instruction, we use a vision-language model (VLM) to give reward feedback to the agent for that task.

In our experiments, we provide two main empirical results.
First, we evaluate our approach in the challenging MineDojo environment.
Here, the agents need to acquire skills that are useful in the environment, such as milking a cow or shearing a sheep in a scene with a cow or sheep.
We demonstrate that conventional unsupervised RL techniques fail since maximizing diversity in complex environments is not sufficient to direct exploration toward useful skills.
In contrast, our approach acquires semantic guidance from foundation models and thus learns semantically meaningful skills.
Second, our empirical analysis highlights the shortcomings of current VLMs for reward computation. 
We further discuss several strategies to improve reward quality, e.g., via policy initialization and reward post-processing.
Our analysis results support that the quality of the VLM-based reward function is limiting, thereby compelling the use of pre-trained behavior priors or reward post-processing.

To summarize, our contributions are as follows:
\begin{itemize}
    \item We present \our,
    unsupervised RL approach that leverages foundation models to guide agent training,
    \item We verify the efficacy of \our in a challenging open-ended environment 
    and demonstrate that our approach outperforms prior unsupervised RL approaches.
    \item We perform extensive qualitative analysis, including learned behavior quality and limitations from foundation model components, to inform future research.
\end{itemize}

    \section{Problem Formulation}
Our goal is to train a policy from a given environment without human-supervised reward signals, similar to the formulation of unsupervised RL~\cite{DIAYN,APT,CIC,MaxEnt,DADS,LSD}. We use foundation models (FMs), i.e., large language models (LLMs) and vision-language models (VLMs), to achieve this goal. 
While our FMs have received much human supervision through content on the internet created and annotated by humans, this content has been distilled into the weights of the FMs and therefore comes ``for free'' during the training of the RL agent. 

Formally speaking, given an environment defined by reward-free Markovian decision process (MDP) $\mathcal{M} = \left( \mathcal{S}, \mathcal{A}, \mathcal{P}, \rho \right)$ with state space $\mathcal{S}$, action space $\mathcal{A}$, transition probability $\mathcal{P}$, and an initial state distribution $\rho$, we aim to pre-train a policy $\pi$, receiving the observation $o \in \mathcal{O}$ in a partially observable MDP $\hat{\mathcal{M}} = \left( \mathcal{S}, \mathcal{O}, \mathcal{A}, \mathcal{P}, \rho \right)$, which can quickly acquire target tasks representing a set of meaningful skills, sampled from unknown task distribution. 
In this work, we focus on zero-shot adaptation given a language instruction, thus each task also contains a language instruction $\delta$ describing the goal.

We assume that the agents have access to a text description of the interactable objects available in the environment $d(s_0)$ with $s_0 \sim \rho$.
As our key idea is to leverage the semantic understanding of foundation models, the description is used to ground a text-based foundation model to the current state of the environment. 
We note that this assumption is more practical than the assumptions in prior works where language descriptions of the full agent state or trajectories were assumed~\cite{ELLM,ALFWorld}, as obtaining an object list is easily implementable in the real world with a pre-trained object detector~\citep{gu2022openvocabulary}.

\section{\our Framework}

We present \ourlong, a closed-loop system for unsupervised RL with foundation model guidance.
Our framework consists of two phases: \textbf{1) LLM task instruction proposal} which obtains a set of semantically meaningful instruction proposals based on the current training environment, and \textbf{2) VLM-guided policy learning} which trains a multi-task policy in the training environment using the obtained instructions, as illustrated in \cref{fig:overview}.
We describe how the agents acquire meaningful instruction proposals grounded in the current environment in \cref{sec:task_proposal}, how we pre-train a policy based on the proposed instructions in \cref{sec:rewarding}, and the practical implementation of our framework in \cref{sec:implementation}.

\subsection{LLM Task Instruction Proposal}\label{sec:task_proposal}

\begin{wrapfigure}[16]{R}{0.48\textwidth}
\vspace{-1.1cm}
  \begin{center}
    \begin{tcolorbox}
        \indent Propose a skill for an Minecraft agent.\newline
        \newline
        You see blocks of \textcolor{red}{grass, grass block}.\newline
        You face entities of \textcolor{blue}{cow}.\newline
        You have a \textcolor{orange}{bucket} in your hand.\newline
        Target Skill: milk a cow.\newline
        \newline
        You see blocks of \textcolor{red}{[objects]}.\newline
        You face entities of \textcolor{blue}{[entities]}. \newline
        You have a \textcolor{orange}{[item]} in your hand.\newline
        Target Skill:
    \end{tcolorbox}
  \end{center}
  \caption{\textbf{An example LLM prompt.} See \cref{app:prompt} 
 for the full prompt and example results.}
 \label{fig:prompt}
\end{wrapfigure}
The first step of our approach is to generate a set of~\emph{imagined} task instructions that are useful to accomplish future target tasks.
To achieve this without human supervision, the agent initialized in a training environment asks an off-the-shelf LLM to generate useful task instructions given the description of the available objects in the current environment $d(s_0)$.
For example, by prompting the observation of a cow and a bucket in the current environment, the LLM suggests the instruction of ``milk a cow.''
The initial state and the proposed task instructions are then used to train a multi-task language-conditioned policy, as described in the next section.
For an example of the prompt we use, see \Cref{fig:prompt}.

\subsection{VLM-Guided Policy Learning}\label{sec:rewarding}
Given the proposed set of $N$-numbers of task instructions $\{\delta^{(i)}\}_{i=1}^N$ and their corresponding initial states,  
our goal is to train a multi-task policy $\pi(a|s, \delta)$ that follows the instructions.
To accomplish this without expert demonstrations or human reward engineering, we leverage CLIP4Clip~\cite{CLIP4CLIP}-style VLMs, which are pre-trained to align videos and corresponding language labels on large video-language datasets, e.g., YouTube data, with contrastive learning. 
We use these VLMs to directly reward the RL agent based on the task instructions and videos of corresponding agent behavior. 

Analogous to the training scheme in MineDojo \cite{MineDojo}, our policy is trained to follow the given instruction by maximizing the VLM ``alignment score'' between the current observation and the instruction as its reward.
Specifically, our reward is defined by:
\begin{equation}
     r_t = r(o_{t-H:t}, \delta) = \frac {
        \phi_V(o_{t-H:t})^\top \phi_T(\delta)
     } {
        |\phi_V(o_{t-H:t})| \cdot |\phi_T(\delta)|
     },
\label{eq:reward}
\end{equation}
where $o_t$ is the visual observation of time step $t$ with $o_{t-H:t}$ implying the sequence of observations with size of $H$, $|\cdot|$ refers to L2-norm of a vector, $\phi_T$ and $\phi_V$ are the text and video encoder of the VLM, $\delta$ is the language instruction, and $H$ is the length of video that the VLM takes. 
This reward scheme works because video-language alignment VLMs are pre-trained to maximize the cosine similarity of correctly paired videos and language instructions, which is directly represented in Equation~\ref{eq:reward}.

Additionally, we empirically observe that the computed reward signal typically exhibits high variance, likely coming from noise induced by visual variations within the observed agent trajectories.
Thus, we adapt the reward stabilization technique from DECKARD~\cite{DECKARD} to mitigate learning instability: 
\begin{equation}
\begin{aligned}
    \tilde{r}_t &= a \cdot \textrm{max}(0, \textrm{avg}(r_{t-W:t}) - b) \\
    \hat{r}_t &= \begin{cases}
        \tilde{r}_t, &\text{if}\; \tilde{r}_t > \textrm{max} (\tilde{r}_{:t-1}) \\
        0,           &\text{otherwise}
    \end{cases}
\end{aligned}
\label{eq:reward_smoothing}
\end{equation}
where avg$(\cdot)$ is the average function, $a$ is a reward scaling factor, $b$ is a bias used for reward shaping, and $W$ is the size of a softening window length.
The importance of this reward post-processing for learning success is discussed in Section~\ref{sec:analysis}. 

Finally, we train a language-conditioned multi-task policy $\pi$ by optimizing the following multi-task objective using any standard RL algorithm: 
\begin{equation}
\scalebox{0.98}{$
    \sum_{i=1}^{N} \mathbb{E}_{o_t \sim \pi, \rho, \mathcal{P}} \left[
      \sum_{t} {\hat{r}(o_{t-H:t}, \delta^{(i)})}
    \right]
$}.
\label{eq:rl_objective}
\end{equation}

\subsection{Practical Implementation}\label{sec:implementation}
\textbf{Foundation Models}
We use Vicuna-13B~\citep{vicuna} for the task proposal and MineCLIP~\citep{MineDojo} for the reward-signal VLM without fine-tuning.

\textbf{Task Instruction Proposal}
To amortize the querying cost of the LLM, we initialize the agent in a fixed set of starting states (i.e., individual Minecraft environments) and pre-compute possible task instructions using the LLM in these scenarios. 

\textbf{Policy and Value Network}
Our policy network is initialized with VPT~\cite{VPT}, a pre-trained behavior cloning policy in the Minecraft domain.
A randomly initialized policy is not able to learn meaningful behavior mainly due to the complexity of the environment and reward quality (see~\cref{sec:analysis}).
To share the pre-trained state representation from VPT, our value function is implemented by adding a single fully-connected layer after the transformer layers.

To fine-tune the pre-trained policy and value function networks into multi-task language-conditioned networks, we add task-conditioning adapter layers~\cite{TransformerAdaptor} in the transformer layers of VPT.
Each adapter layer concatenates an embedding of language instruction for a current task with an output from the previous layer.
The concatenated vector is then used to compute the residual prediction of the previous output as proposed in~\cite{TransformerAdaptor}.
We only update the parameters of the adapter layers and a value function head, while other parameters are frozen.
See~\cref{app:training} for more details of the policy architecture.

\textbf{RL Optimization}
Our policy is trained with the PPO algorithm \cite{PPO}.
A PPO iteration uses one rollout for each proposed task to optimize the multi-task objective in~\cref{eq:rl_objective}.
Following~\cite{VPT}, the KL-divergence term is added to~\cref{eq:rl_objective} to prevent catastrophic forgetting.
We provide hyperparameters for optimization in~\cref{app:training}.
    \section{Experiments}
Our experiments aim to tackle two questions:
(1) Can the agents acquire meaningful behaviors in a challenging, open-ended environment using the \our framework? and 
(2) Do current off-the-shelf foundation models demonstrate sufficient understanding of agent behaviors to enable scalable learning within the \our framework?
\cref{sec:result} presents a comparison experiment against baselines answering the first question, and \cref{sec:analysis} highlights the reward quality from VLM with respect to the second question. 

\subsection{Experimental Setup}\label{sec:exp_setup}

We evaluate our method with other baselines in MineDojo~\cite{MineDojo} environment.
Minedojo is an open-ended environment with a variety of potential open-ended goals in simulated 3D game worlds.
In our experiments, the agents perceive the world through RGB images and interact with the environment by choosing one of the discrete actions for each time step.

In particular, we choose 8 different environment setups from a task set proposed in \cite{MineDojo}.
Each environment setup is designed to evaluate a basic skill in Minecraft game, such as harvesting logs or combating monsters, thus necessary objects or animals are placed near the initial position or a necessary tool is provided to the agent.

For the task proposal, as described in \cref{sec:task_proposal}, we prompt an LLM with the initial textual information of nearby objects, entities, and the current tool in the agent's hand retrieved from the game simulator state.

For the training, we use 5 different world seeds from each environment setup, resulting in 40 different initial states.
The initial state is used for the task proposal, thus the number of LLM-generated task instructions is also 40.
Note that the agent is not provided with any supervision about the evaluation tasks, such as the oracle task instruction or reward, for training.

After the training, we zero-shot evaluate the learned policy.
We rollout the learned policy 20 times for each environment setup and measured the average success rates with the success criteria of each evaluation task.
The language instruction for each evaluation task is provided to the policy if the policy is conditioned on language.
For more details, see \cref{app:experiment}.

We compare our method with below baselines:
\begin{itemize}[leftmargin=0.2in]
    \item[] \textbf{\our w/ Oracle Tasks} is trained with the reward computed by the definition in \cref{eq:reward}, similar to \our, but directly uses the oracle task instruction annotated by humans.
    \item[] \textbf{APT} maximizes the state-entropy in visual embedding space by using variational autoencoder (VAE)~\cite{VAE}. The VAE is pre-trained with visual observations from VPT policy rollouts.
    \item[] \textbf{APT w/ MineCLIP} is similar with \textbf{APT}, but maximizes the state-entropy in visual embedding space of MineCLIP, instead of VAE.
    \item[] \textbf{VPT} is directly evaluated with the pretrained VPT \cite{VPT} policy and is not finetuned. This baseline indicates the lower bound of our framework. 
\end{itemize}

All methods are trained for 300 iterations of PPO update in the same training environment.
For a fair comparison, all baselines except VPT use VPT-initialized policy and adapter layers for fine-tuning.
The trajectories used for plotting and computing the reward scores are with 10 rollouts in a fixed evaluation task with the predefined spawn region of the target entity, and the scores are normalized with the maximum and minimum values of the whole collection.
For ablation studies in \cref{sec:analysis}, we train for 100 iterations of PPO updates.
More details on the baseline implementations are in~\cref{app:baseline}.

\begin{figure}[t]
  \begin{minipage}[h]{0.33\textwidth}
    \centering
    \vspace{0.3cm}
    \scalebox{0.75}{
        \renewcommand{\arraystretch}{1.2}
    
\begin{tabular}{c | c}
    \toprule
        & Success Rate \\ 
    \midrule
\multilinecell{VPT}
        & \multilinecell{0.19 \scriptsize{$\pm$ 0.01}} \\
    \midrule
\multilinecell{APT}
        & \multilinecell{0.14 \scriptsize{$\pm$ 0.02}} \\
\multilinecell{APT w/ MineCLIP}   
        &  \multilinecell{0.16 \scriptsize{$\pm$ 0.01}} \\
    \midrule 
\multilinecell{\our}
        & \multilinecell{\textbf{0.50 \scriptsize{$\pm$ 0.03}}} \\  
    \midrule
\multilinecell{\our w/ Oracle Tasks}
        & \multilinecell{0.49 \scriptsize{$\pm$ 0.02}} \\   
    \toprule
    \end{tabular}
    }
    \vspace{0.2cm}
    \captionof{table}{
        \textbf{Evaluation results.}
        The LiFT agent shows results that outperforms the unsupervised RL methods and is comparable to the baseline using the oracle task instructions annotated by humans.
    }
   \label{tab:successrate}
   \end{minipage}
  \hfill
  \begin{minipage}[h]{0.65\textwidth}
    \scalebox{1.0}{
        \input{figures/qualitative.tex}
    }
    \vspace{-0.1cm}
    \captionof{figure}{
    \textbf{Qualitative results.} 
    The LiFT agent learns how to interact with the target in environment (green). The baselines optimize their own intrinsic reward, resulting in diverse but not semantically meaningful behaviors.
    }
    \label{fig:qualitative}
  \end{minipage}
\end{figure}

\subsection{Results}\label{sec:result}

We present our quantitative comparison results in \cref{tab:successrate} and qualitative examples in \cref{fig:qualitative}.
Our approach outperforms the unsupervised RL baselines (APT and APT w/ MineCLIP) and achieves comparable results to the oracle approach of using human supervision for task proposal (\our w/ Oracle) with zero-shot evaluation.

During the unsupervised pre-training stage, our method succeeds in training a policy equipped with semantically meaningful behavior.
For example, when the agent is placed in an initialization position near a sheep \iconsheep with a shear \iconshear, our framework decides to learn how to shear a sheep.
\cref{fig:qualitative}a shows how our agents interact with the target entity given the task instruction.

On the other hand, the success rates of unsupervised RL baselines are even worse than the VPT baseline.
Each agent of the baselines tries maximizing the diversity, as shown with the highest reward scores in~\cref{fig:qualitative}b.
However, the learned behavior is not semantically meaningful in that the agents just wander around a wide range of regions, as shown in~\cref{fig:qualitative}a, or gaze at diverse entities without following the task description.

\begin{figure}
  \hfill
  \begin{minipage}[h]{1\textwidth}
    \begin{subfigure}[t]{0.494\textwidth}
        \includegraphics[width=\textwidth]{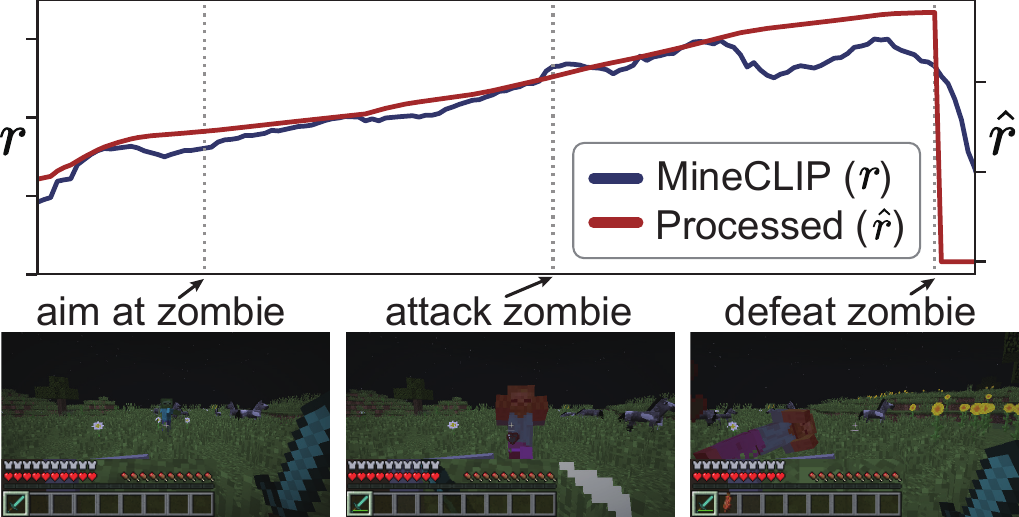}
        \caption{combat a zombie}\label{fig:reward1}
    \end{subfigure}
    \hfill
    \begin{subfigure}[t]{0.494\textwidth}
        \includegraphics[width=\textwidth]{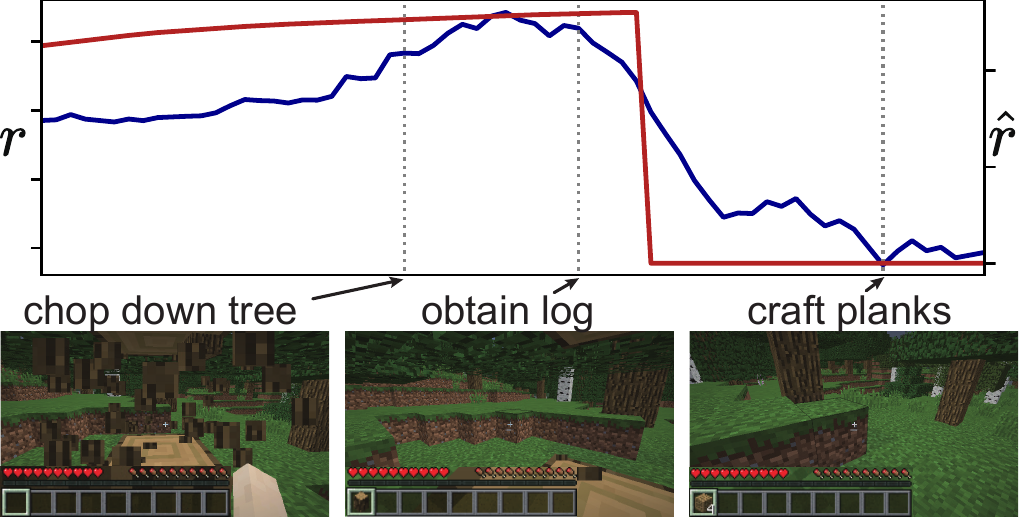}
        \caption{craft planks}\label{fig:reward2}
    \end{subfigure}
  \caption{
  \textbf{VLM reward examples.}
  \ref{fig:reward1} and \ref{fig:reward2} plot computed rewards for each task instruction and its demonstration. While reward stabilization greatly reduces noise, the video-text alignment of VLM is not as precise as providing the highest reward for the most critical behavior in the demonstration.
  }\label{fig:analysis}
  \end{minipage}
\end{figure}

\subsection{Analysis}\label{sec:analysis}

\begin{wrapfigure}{r}{6.5cm}
  \begin{minipage}[h]{0.45\textwidth}
    \vspace{-0.4cm}
    \centering
    \scalebox{0.75}{
    \renewcommand{\arraystretch}{1.2}

\begin{tabular}{c  c  c | c}
    \toprule
      \multilinecell{Reward\\Stabilization} & 
      \multilinecell{Policy\\Initialization} & 
      \multilinecell{Reward\\Type} &
      \multilinecell{Success\\Rate} \\ 
    \midrule
    
         \multilinecell{\textbf{\checkmark}} 
        & \multilinecell{\textbf{\checkmark}}
        & \multilinecell{\textbf{Cossim.}}
        & \multilinecell{\textbf{0.42 \scriptsize{$\pm$ 0.01}}} \\ 
         \multilinecell{} 
        & \multilinecell{\checkmark}
        & \multilinecell{Cossim.}
        & \multilinecell{0.33 \scriptsize{$\pm$ 0.02}} \\  
         \multilinecell{\checkmark} 
        & \multilinecell{}
        & \multilinecell{Cossim.}
        & \multilinecell{0.00 \scriptsize{$\pm$ 0.00}} \\ 
         \multilinecell{\checkmark} 
        & \multilinecell{\checkmark}
        & \multilinecell{Softmax}
        & \multilinecell{0.35 \scriptsize{$\pm$ 0.03}} \\ 
    \toprule
\end{tabular}

    
         

    }
    \vspace{-0.1cm}
    \captionof{table}{\textbf{Ablation results.} The choice of rewarding scheme and VPT initialization are crucial for successful learning.}
    \vspace{-.8cm}
  \label{tab:ablation}
  \end{minipage} 
\end{wrapfigure}

We empirically observe that the main challenge of our framework for scalable learning is obtaining a decent quality of rewards.
To highlight how different design choices influence reward function quality, we present an ablation experiment in~\cref{tab:ablation} and a qualitative analysis of VLM rewards in~\cref{fig:analysis}.

\paragraph{Reward Stabilization}

The reward post-processing in~\cref{eq:reward_smoothing} is crucial for successful learning of our framework, as shown with the comparison between the first row and the second row of~\cref{tab:ablation}.
The examples in \cref{fig:analysis} qualitatively show that post-processing greatly stabilizes the noisy VLM reward, justifying the performance improvement.
While the rewards before processing contain fluctuation from visual changes in observation, the smoothing and clipping in the reward processing stably cancel out the task-irrelevant noise of the original rewards.

\paragraph{Policy Initialization}

The agents are unable to acquire meaningful behaviors in the absence of the VPT policy initialization, as demonstrated in the comparison between the first row and the third row of~\cref{tab:ablation}.
We suspect that this failure primarily stems from the challenge of attempting to learn long-horizon and high-dimensional Minecraft skills entirely from scratch.
However, the results also suggest that the VLM reward lacks the necessary density or precision for efficient training of agents from scratch, further discussed in the next paragraph.

\paragraph{VLM Reward Limitation}

We visualize examples of rewards computed by VLM for  task accomplishing demonstrations in \cref{fig:analysis}.
Both examples in~\cref{fig:analysis} illustrate that VLM rewards are generally high for behavior that is somewhat related to the task instructions but often fall short of providing the highest rewards for the most success-critical moment.

In particular, the agents struggle to learn the full task behavior from the rewards in~\cref{fig:reward2}, as they are dominantly rewarded only for a subtask of the target task.
We suspect that this imperfection arises from the video-language alignment issue with the current off-the-shelf VLM.
This work has attempted to alleviate the issues by using several strategies discussed above, yet we leave improving VLM reward for a more diverse set of tasks, including complex and delicate tasks, as future work.

For further analysis of the limitation of VLMs, we point to the limitations of the internet-scale dataset used for finetuning current off-the-shelf VLMs.
Many of the YouTube videos used for training MineCLIP have misalignment issues between the speech caption used as the text label and the video itself.
For example, the auditory explanation for certain actions comes far after what happens in the scene in many YouTube videos. 
We anticipate that this challenge can be mitigated through fine-tuning with in-domain expert data or data curation of videos with better alignment quality.

Another hypothesis we posit is that the contrastive learning objective without considering \emph{precise success moment} might have not been appropriate for fine-grained feedback on motion.
Similar observations have been reported with video-language alignment score of R3M on robot manipulation tasks \cite{LAMP}. 
We believe that a training objective more suitable for the fine-grained temporal understanding of video-language alignment is required. 

\paragraph{Reward Type}

To examine whether a different formulation of reward can mitigate the aforementioned reward imperfectness, 
we explore a reward type of softmax, proposed in~\cite{MineDojo}.
This exploration is aimed at assessing the potential of relative normalization through softmax rewards for reward stabilization.
For each sample in training batch, we use its original task instruction as a positive instruction and consider the other instructions in the batch as negative instructions to compute the softmax reward.
Given the comparison between the first row and the last row of~\cref{tab:ablation}, we conclude the our method does not significantly benefit from a naive application of softmax reward. 
We leave the careful design of negative instructions as future work.

    \section{Related Work}

\paragraph{Unsupervised Reinforcement Learning}

Unsupervised RL aims to learn a policy, that is helpful to adapt downstream tasks, without human supervision.
While diverse categories of intrinsic rewarding methods are proposed, including information gain~\cite{burda2018exploration,pathak2017curiosity}, competence~\cite{DIAYN,CIC,Sharma2020Dynamics-Aware,Hansen2020Fast,liu2021aps}, or state visitation diversity~\cite{APT,yarats2021reinforcement}, a core idea of these methods is maximizing behavior diversity in a learned representation space without semantic guidance.
As the learned representation is not guaranteed to be well-aligned with human environment understanding, maximizing diversity in the representation spaces often lacks semantic meaningfulness.
In this work, we overcome this limitation with guidance from foundation models.

\paragraph{Open-Ended Learning}

As an effort to develop generally capable agents, recent works have proposed training agents in an open-ended environment instead of a specified task scenario~\cite{OpenMaze, NetHack, BabyAI, MineDojo}.
MineDojo~\cite{MineDojo} casts Minecraft game as a developmental open-ended environment inspired by its high degree of freedom and further suggests to use of internet-scale knowledge about the game for open-ended learning.
While MineDojo proposes a set of new tasks brainstormed with LLM and to reward agents using VLM trained on the YouTube knowledge base, how to integrate and ground them in an open-ended learning environment is yet explored.
Voyager~\cite{Voyager} introduces a lifelong learning framework that continually explores the world and extends the skill set in the MineDojo environment using LLM.
Voyager uses LLM for code-level planning and refinement based on pre-defined basic skills, while our work aims to acquire visuo-motor policy without access to hand-engineered skills.

\paragraph{Foundation Models for RL}

The internet-scale knowledge capacity of recent FMs enables automating impractical human effort in RL framework~\cite{rt22023arxiv,DECKARD,LACO,VoxPoser,CompFM,LAMP}.
Existing works query pre-trained LLMs for tasks to learn~\cite{ELLM,OMNI,MineDojo}, language-level plans~\cite{progprompt}, and language labels~\cite{zhang2023bootstrap,zhang2023sprint}; or use pre-trained VLMs to obtain visual feedback~\cite{ZEST,DECKARD,R3M,roboclip2023}.
Closest to ours, ELLM~\cite{ELLM} uses LLMs to propose new tasks for agents to learn. 
However, their experiments are limited to simpler environments which a captioning model can generate meaningful captions for, as they use generated caption alignment with proposed tasks to reward the agent. 
A line of work~\cite{DECKARD,MineDojo,Voyager, Steve-1} specifically focuses on using FMs for the Minecraft domain, while none of the works integrate pre-trained LLM and VLM for an unsupervised RL system.
We focus on closing the loop for adaptation to open-ended environments.
    \section{Conclusion}

We propose \our, an unsupervised RL framework for learning semantically meaningful behavior without human supervision.
Our method uses two foundation models, LLM and VLM, to construct a closed-loop policy learning system.
Our experimental results in the MineDojo environment demonstrate that agents with \our framework can acquire semantically meaningful behavior in a challenging open-ended environment.
We demonstrate analysis for our strategies and the quality of the learned behaviors, showing the limitation of the prior approaches in unsupervised RL that we compare with.
Our further analysis reveals imperfections in the quality of VLM rewards, we hope future work on enhancing VLM to provide higher-quality visual feedback.

    \bibliographystyle{unsrtnat}
    \bibliography{bibtex}
    \newpage
    \appendix
\section*{Supplementary Material}
\section{Experiment Detail}\label{app:experiment}

\paragraph{Observation and Action space}
In our experiment, the agents observe $640 \times 360$ RGB rendered images of the Minecraft game.
For action space, we follow the hierarchical action space of~\cite{MineDojo} to incorporate with a complex control interface of Minecraft game.

\paragraph{Training} 
For training, the environment for training is in the total number of 40 with 3 different seeds. 
We select 8 different environment setups, sharing the biome, candidate spawning entity, and time.
Among 8 environment setups, the biomes of 7 environment setups are chosen to be `sunflower plains' where candidate target entities can be such as cow, chicken, sheep, horse, and so forth for the day and spider, zombie, and so on.
For the other worlds, the biomes are chosen to be `forest', where it is with more trees around with possible candidates similar to `sunflower plains'.
We select 6 environment setups to be in the day and 2 environment setups to be in the night. 

Then, we set 5 different world seeds for each environment setup, so 40 different environments in total.
The below \cref{tab:seed} shows the 5 different world seeds used for each setup.
\begin{table}[h!]
\begin{center}
\scalebox{0.9}{
\begin{tabular}{c | c }
\toprule
       seed & world seeds \\ 
    \midrule
1
& 171529 167223 177048 159010 148055
\\

2
& 189255 172172 195815 144565 131018
\\

3
& 168267 143566 142612 145890 121242
\\
\toprule
\end{tabular}
}
\caption{Table of values for world seeds.}
\vspace{-0.8cm}
\label{tab:seed}
\end{center}
\end{table}

For easier skill learning, we initialize a possible target entity in front of the agent for each environment.
The spawn region is to be from $[-4, 1, 5]$ to $[4, 1, 8]$ in xyz-coordinate.
The entities are randomly spawned in the spawn region but do not necessarily become the target of the task.

For training, we run 300 epochs of PPO step for each baseline.
We set the time limit to 200 environment steps for each training epoch.


\paragraph{Evaluation}
For evaluation, the agents are asked to perform 8 tasks. We modify action space and environment, as explained in detail below.
In the evaluation, the agents are asked to perform target tasks described in textual form. For one evaluation, we rollout for 20 times and compute the average success rate for the comparison results. The evaluation scores are measured when the training of 260, 280, and 300 epochs is over for each baseline. We report the average of three measures and the standard error of them with respect to three seeds.

We use random values for the world seeds to be different from the world seeds used in training. We exhibit the target tasks with human-annotated task descriptions as below.

\begin{table}[h!]
\begin{center}
\scalebox{0.9}{
\begin{tabular}{c | c c c | c }
\toprule
        & target entity & item & task description \\ 
    \midrule
1
& cow
& bucket
& get milk from a cow
\\

2
& sheep
& shear
& shear a sheep and get some wool
\\

3
& chicken
& diamond sword
& hunt a chicken and get its meat
\\

4
& tree
& (nothing)
& collect logs
\\

5
& cow
& diamond sword
& kill a cow
\\

6
& sheep
& diamond sword
& kill a sheep
\\

7
& spider
& diamond sword
& kill a spider
\\

8
& zombie
& diamond sword
& kill a zombie
\\
\toprule
\end{tabular}
}
\caption{Table of information on the target task environment.}
\vspace{-0.6cm}
\label{tab:target_task}
\end{center}
\end{table}

The tasks are chosen to be diverse enough and meaningful for initial survival in the Minecraft environment from a task set proposed in \cite{MineDojo}.

For the qualitative analysis, we test the evaluation of the task `shear a sheep and get some wool'. We sample 10 trajectories for each model we test.

\newpage
\section{Full Prompt}\label{app:prompt}
The full prompt we use is as below. We use lidar rays in the MineDojo environment, with the range of from $[-45\pi / 180, -45\pi / 180, 20]$ to $[+45\pi / 180, +45\pi / 180, 20]$, for obtaining the information of initial surroundings. An agent gets the information of objects, and entities in front of it and an item in its hand. This information is parsed into the prompt.

\begin{center}

    \begin{tcolorbox}
    Below is an instruction that describes a task, paired with an input that provides further context.
    Write a response that appropriately completes the request.\newline
    \newline
    Instruction:\newline
    You are proposing basic skills that a Minecraft agent can learn.\newline
    You just began Minecraft game.\newline
    Propose one of the most interesting tasks that you can learn from a given situation in the input as `Target Task'.\newline
    Propose several not interesting tasks as `Negative Tasks'.\newline
    \newline
    Guidelines:\newline
    Prioritize using your item in your hand and target entities you face.\newline
    And list other possible tasks that are possible but less interesting tasks.\newline
    Each task should refer to one specific object in the input.\newline
    Make sure that the response is as concise as possible, like a pair of an action and a target object.\newline
    Answer with each task of less than 5 words.\newline
    \newline
    ----------------------------------------------------------\newline
    Input:\newline
    You see blocks of grass, grass block, wheat, wood, flower.\newline
    You face entities of horse.\newline
    You have a diamond sword in your hand.\newline
    Response:\newline
    Target Task: combat a horse.\newline
    Negative Tasks: dig the ground, water the flower, grow a plant.\newline
    \newline
    ----------------------------------------------------------\newline
    Input:\newline
    You see objects of \{(', '.join(objects)) if len(objects) > 0 else 'nothing'\}.\newline
    You face entities of \{(', '.join(entities)) if len(entities) > 0 else 'nothing'\}.\newline
    You have \{item\} in your hand.\newline
    \newline
    Response:\newline
    Target Task: 
    \end{tcolorbox}

\end{center}

\newpage
We also provide a sample set of results, in the table below. We leave (nothing) when the agent does not observe anything or does not have anything in its hand. The observation heavily depends on stochasticity. Small entities like chicken, sometimes, do not get detected by the lidar, although it's observable in the screenshot. We observe that the LLM may propose (nothing) as a target task.
\begin{table}[h!]
\begin{center}

\scalebox{0.8}{
\begin{tabular}{c | c  c  c | c}
    \toprule
       & objects & entities & item & task instruction proposal \\ 
    \midrule
    
1
& flower grass grass block plant 
& cow sheep 
& bucket
& milk entities
\\

2
& grass grass block plant 
& (nothing)
& bucket
& watering plants
\\

3
& dirt flower grass grass block water 
& cow 
& bucket
& water the flower
\\

4
& dirt grass grass block plant 
& cow 
& bucket
& milk a cow
\\

5
& flower grass grass block plant 
& cow sheep 
& bucket
& water the cow/sheep
\\

6
& flower grass grass block plant water 
& chicken sheep 
& shears
& Shear the sheep
\\

7
& grass grass block plant 
& sheep 
& shears
& shear a sheep
\\

8
& grass grass block plant 
& (nothing)
& shears
& cut grass
\\

9
& dirt flower grass grass block plant 
& sheep 
& shears
& shear a sheep
\\

10
& flower grass grass block plant 
& pig sheep 
& shears
& shear animals
\\

11
& flower grass grass block plant water 
& (nothing)
& (nothing)
& drink water
\\

12
& grass grass block 
& (nothing)
& (nothing)
& observe
\\

13
& plant 
& (nothing)
& (nothing)
& observe plants
\\

14
& dirt grass grass block stone water 
& (nothing)
& (nothing)
& explore
\\

15
& flower grass grass block plant 
& sheep 
& (nothing)
& shepherd sheep
\\

16
& grass grass block leaves wood 
& (nothing)
& (nothing)
& explore
\\

17
& grass grass block leaves wood 
& pig 
& (nothing)
& chase a pig
\\

18
& grass grass block leaves wood 
& (nothing)
& (nothing)
& gather objects
\\

19
& dirt grass grass block leaves stone wood 
& (nothing)
& (nothing)
& collect dirt
\\

20
& dirt grass block leaves wood 
& (nothing)
& (nothing)
& walk towards nearest object
\\

21
& flower grass grass block plant water 
& chicken cow sheep 
& diamond sword
& kill a chicken
\\

22
& grass grass block plant 
& cow 
& diamond sword
& kill cow
\\

23
& plant 
& (nothing)
& diamond sword
& harvest plant
\\

24
& dirt grass grass block plant 
& cow 
& diamond sword
& attack cow
\\

25
& grass grass block plant 
& cow pig sheep 
& diamond sword
& hunt an animal
\\

26
& flower grass grass block plant water 
& chicken sheep 
& diamond sword
& herd animals
\\

27
& grass grass block plant 
& sheep 
& diamond sword
& shear sheep
\\

28
& dirt flower grass grass block plant water 
& (nothing)
& diamond sword
& kill a plant
\\

29
& dirt grass grass block plant 
& sheep 
& diamond sword
& kill a sheep
\\

30
& grass grass block plant 
& sheep 
& diamond sword
& shear a sheep
\\

31
& flower grass grass block plant water 
& sheep spider 
& diamond sword
& kill a spider
\\

32
& grass grass block plant 
& spider 
& diamond sword
& kill spider
\\

33
& grass grass block plant 
& (nothing)
& diamond sword
& (nothing)
\\

34
& dirt flower grass grass block plant 
& spider 
& diamond sword
& kill spider
\\

35
& grass grass block plant 
& sheep spider 
& diamond sword
& kill spider
\\

36
& flower grass grass block plant water 
& chicken zombie 
& diamond sword
& combat a zombie
\\

37
& grass grass block plant 
& zombie 
& diamond sword
& combat zombie
\\

38
& grass grass block plant 
& (nothing)
& diamond sword
& Harvest grass
\\

39
& dirt flower grass grass block plant 
& zombie 
& diamond sword
& slay zombie
\\

40
& grass grass block plant 
& zombie 
& diamond sword
& kill the zombie
\\
\toprule
\end{tabular}
}
\caption{A result of task instruction proposal from LLM}
\label{tab:answer}
\end{center}
\end{table}

\newpage
\section{Training Detail}\label{app:training}

\paragraph{Network Architecture}
As described in~\cref{sec:implementation}, we added multi-task adapter layers in the transformer layers of VPT policy.
Specifically, we added adapter layers after every feed-forward network in transformer layers, and the adapter output is computed by:

\begin{equation}
    \mathbf{h'} = \mathbf{h} + g(\mathrm{ReLU}(f(\mathbf{h} || \phi_T(\delta)))),
\end{equation}

where $||$ represents a vector concatenation, $\mathbf{h}$ is an output from the previous feed-forward network, $\delta$ is a task instruction of current observation, f is a down-projecting linear layer, and g is an up-projecting linear layer.
We used publicly opened \texttt{bc-house-3x} VPT architecture and the pre-trained weights.
We used MineCLIP as our VLM for the Minecraft domain and \texttt{mineclip\_attn} variant if used for our experiments.

\paragraph{Hyperparameters} We provide a hyperparameter set for our training: 
\begin{center}
    \begin{tabular}{c | c }
    \toprule
    name & value \\ 
    \midrule
    PPO clip & 0.2 \\
    PPO learning rate & 0.0001 \\
    PPO max epoch & 5 \\
    PPO batch size & 80 \\
    PPO target KL & 0.1 \\
    Value function loss scale & 1 \\
    $\gamma$ & 0.999 \\
    GAE $\lambda$ & 0.95 \\
    GAE normalized & True \\
    Adapter contraction factor & 8 \\
    VPT KL loss scale & 0.3 \\ 
    VPT KL loss scale decay & 0.999 \\
    Eq.\ref{eq:reward} $H$ & 16 \\
    Eq.\ref{eq:reward_smoothing} $a$ & 0.005 \\
    Eq.\ref{eq:reward_smoothing} $b$ & 21 \\
    Eq.\ref{eq:reward_smoothing} $W$ & 50 \\
    \bottomrule
    \end{tabular}
\end{center}
\captionof{table}{\textbf{Hyperparameters for \our}}

\paragraph{Action Space Mapping} 
As VPT is pre-trained on~\cite{minerl} environment, the action space of Minedojo and VPT policy output has a discrepancy.
We mapped the inventory opening action of VPT into the crafting action of Minecraft action space to enable our VPT-based policy to craft items.
We introduced an additional policy head for a categorical distribution over a set of craftable items in the Minedojo environment, where we randomly initialized the new head.

\newpage
\section{Baseline Implementation}\label{app:baseline}
We explain details of objectives and hyperparameters used during the experiment for the baselines.

\paragraph{\our w/ Oracle Tasks}
For `\our w/ Oracle Tasks' baseline, we used human-annotated task description for training described in \cref{tab:target_task}, without asking LLM for task instruction proposal.

\paragraph{APT, APT w/ MineCLIP}
For unsupervised RL baselines, we compute intrinsic rewards for particle-based entropy maximization in representation space following~\cite{APT}.

$\newline$
For `APT' baseline, we first pre-trained a VAE on a set of VPT rollouts and used its encoder for the agent behavior learning.
We used an image encoder of VAE as and representation encoder, instead of an encoder trained by contrastive loss, for simplicity.
We also observed that continually fine-tuning the representation encoder during the agent behavior learning does not help to learn useful representation but rather introduces large training instability.

\begin{center}
    \begin{tabular}{c | c }
    \toprule
    name & value \\ 
    \midrule
    Buffer size & 80000 \\
    Batch size & 64 \\
    Number of particles & 4096 \\
    $k$ & 96 \\
    Scale & 0.01 \\
    \bottomrule
    \end{tabular}
\captionof{table}
{\textbf{Hyperparameters for APT}}
\end{center}

$\newline$
For VAE architecture used in `APT' baseline, we let a sequence of ResNet\cite{ResNet} blocks for both encoder and decoder. One ResNet block is composed of two ResNet layers with batch normalization after each layer. We define the reconstruction objective with mean-squared error loss. We resize the observation resolution to feed into the VAE encoder. We provide a hyperparameter set.

\begin{center}
    \begin{tabular}{c | c }
    \toprule
    name & value \\ 
    \midrule
    Input image & 64x64x3 \\
    Latent dimension & 512x4x4 \\
    Encoder blocks channels & (64, 128, 256, 512) \\
    Decoder blocks channels & (512, 256, 128, 64)\\
    Activation function & ELU \\
    VAE learning rate & 5e-4 \\
    VAE learning epoch & 5 \\
    \bottomrule
    \end{tabular}
\end{center}
\captionof{table}{\textbf{Hyperparameters for VAE}}

$\newline$
For the `APT w/ MineCLIP' baseline, We used MineCLIP visual encoder $\phi_T$ instead of the pre-trained VAE to compute the representations of the transitions. \texttt{mineclip\_attn} variant if used for our experiments.

\begin{center}
    \begin{tabular}{c | c }
    \toprule
    name & value \\ 
    \midrule
    Buffer size & 20000 \\
    Batch size & 64 \\
    Number of particles & 1024 \\
    $k$ & 12 \\
    Scale & 0.01 \\
    \bottomrule
    \end{tabular}
\captionof{table}
{\textbf{Hyperparameters for APT w/ MineCLIP}}
\end{center}

\paragraph{VPT} 
For the `VPT' baseline, We used publicly opened \texttt{bc-house-3x} VPT architecture and the pre-trained weights without finetuning during the evaluation.

\end{document}